\documentclass{article}

\usepackage{microtype}
\usepackage{graphicx}
\usepackage{subcaption}
\usepackage{booktabs} 
\usepackage{xcolor}
\usepackage{hyperref}
\usepackage{graphicx}
\usepackage{amsmath}

\usepackage{amssymb}
\usepackage{algorithm,algorithmic}
\usepackage{caption}
\usepackage{tabularx}
\usepackage[accepted]{icml2018}
\icmltitlerunning{Semi-Amortized Variational Autoencoders}
\newcommand{\xvec}{\mathbf{x}}
\newcommand{\vvec}{\mathbf{v}}

\newcommand{\zvec}{\mathbf{z}}

\newcommand{\uvec}{\mathbf{u}}

\newcommand{\hvec}{\mathbf{h}}
\newcommand{\mcL}{\mathcal{L}}
\newcommand{\mcD}{\mathcal{D}}
\newcommand{\mcN}{\mathcal{N}}
\newcommand{\mcU}{\mathcal{U}}
\newcommand{\E}{\mathbb{E}}
\newcommand{\reals}{\mathbb{R}}
\newcommand{\Hess}{\mathrm{H}}
\DeclareMathOperator*{\KL}{KL}
\DeclareMathOperator*{\ELBO}{ELBO}

\DeclareMathOperator*{\MLP}{MLP}
\DeclareMathOperator*{\LSTM}{LSTM}
\DeclareMathOperator*{\softmax}{softmax}
\DeclareMathOperator*{\enc}{enc}

\DeclareMathOperator*{\clip}{clip}
\newcommand*\diff{\mathop{}\!\mathrm{d}}
\newcommand{\given}{\,|\,}
\newcommand{\param}{;}

\begin{document}

\twocolumn[
\icmltitle{Semi-Amortized Variational Autoencoders}




\begin{icmlauthorlist}
\icmlauthor{Yoon Kim}{har}
\icmlauthor{Sam Wiseman}{har}
\icmlauthor{Andrew C. Miller}{har}
\icmlauthor{David Sontag}{mit}
\icmlauthor{Alexander M. Rush}{har}
\end{icmlauthorlist}

\icmlaffiliation{har}{School of Engineering and Applied Sciences, Harvard University, Cambridge, MA, USA}
\icmlaffiliation{mit}{CSAIL \& IMES, Massachusetts Institute of Technology, Cambridge, MA, USA}
\icmlcorrespondingauthor{Yoon Kim}{yoonkim@seas.harvard.edu}

\icmlkeywords{Machine Learning, ICML}

\vskip 0.3in
]



\printAffiliationsAndNotice{}  

\begin{abstract}
Amortized variational inference (AVI) replaces instance-specific local inference with a global inference network. While AVI has enabled efficient training of deep generative models such as variational autoencoders (VAE), recent empirical work suggests that inference networks can produce suboptimal variational parameters. We propose a hybrid approach, to use AVI to initialize the variational parameters and run stochastic variational inference (SVI) to refine them. Crucially, the local SVI procedure is itself differentiable, so 
the inference network and generative model can be trained end-to-end with gradient-based optimization. 
This semi-amortized approach enables the use of rich generative models without experiencing the posterior-collapse phenomenon 
common in training VAEs for problems like text generation. Experiments show this approach outperforms strong autoregressive and variational baselines on standard text and image datasets.
\end{abstract}
\vspace{-8mm}
\section{Introduction}
\vspace{-2mm}
Variational inference (VI) \cite{Jordan1999,Wainwright2008} is a framework 
for approximating an intractable distribution by optimizing over a family of tractable
surrogates. Traditional VI algorithms iterate over the observed data and update the variational parameters with closed-form coordinate ascent  updates that exploit conditional conjugacy  \cite{Ghahramani2001}. This style of optimization is challenging to extend to large datasets and non-conjugate models.
However, recent advances in \emph{stochastic} \cite{Hoffman2013}, \emph{black-box} \cite{Ranganath2014,Ranganath2016}, and \emph{amortized} \cite{Mnih2014,Kingma2014,Rezende2014} variational inference have made it possible to scale to large datasets and rich, non-conjugate models (see \citet{Blei2017}, \citet{Zhang2017} for a review of modern methods).

In \textit{stochastic variational inference} (SVI), the variational parameters for each data point are randomly initialized and then optimized to maximize the evidence lower bound (ELBO) with, for example, gradient ascent. These updates are based on a subset of the data, making it possible to scale the approach. In \textit{amortized variational inference} (AVI), the local variational parameters are instead predicted by an inference (or recognition) network, which is shared 
(i.e. amortized) across the dataset. Variational autoencoders (VAEs) are deep generative models that utilize AVI for inference and jointly train the generative model alongside the inference network.

SVI gives good local (i.e. instance-specific) distributions within the variational family but requires performing optimization for each data point. AVI has fast inference, but having the variational parameters be a parametric function of the input may be too strict of a restriction. As a secondary effect this may militate against learning a good generative model since its parameters may be updated based on suboptimal variational parameters. \citet{Cremer2017} observe that the \emph{amortization gap} (the gap between the log-likelihood and the ELBO due to amortization) can be significant for VAEs, especially on complex datasets.

Recent work has targeted this amortization gap by combining amortized inference with iterative refinement during training \cite{Hjelm2016,Krishnan2017}. These  methods use an encoder to initialize the local variational parameters, and then subsequently run an iterative procedure to refine them. 
To train with this hybrid approach, they utilize a separate training time objective. For example \citet{Hjelm2016} train the inference network to minimize the KL-divergence between the initial and the final variational distributions, while \citet{Krishnan2017} train the inference network with the usual ELBO objective based on the initial variational distribution.

In this work, we address the train/test objective mismatch and consider methods for training \textit{semi-amortized variational autoencoders} (SA-VAE) in a fully end-to-end manner.  We propose an approach that leverages differentiable optimization \cite{Domke2012,Maclaurin2015,Belanger2017} and differentiates \emph{through} SVI while training the inference network/generative model. We find that this method is able to both improve estimation of variational parameters and produce better generative models. 

We apply our approach to train deep generative models of text and images, and observe that they outperform autoregressive/VAE/SVI baselines, in addition to direct baselines that combine VAE with SVI but do not perform end-to-end training. We also find that under our framework, we are able to utilize a powerful generative model without experiencing the ``posterior-collapse'' phenomenon often observed in VAEs, wherein the variational posterior collapses to the prior and the generative model ignores the latent variable \cite{Bowman2016,Chen2017,Zhao2017}. This problem has particularly made it very difficult to utilize VAEs for text, an important open issue in the field. With SA-VAE, we are able to outperform an LSTM language model by utilizing an LSTM generative model that maintains non-trivial latent representations. Code is available at \url{https://github.com/harvardnlp/sa-vae}.
\vspace{-3mm}
\section{Background}
\vspace{-1mm}
\paragraph{Notation}
Let $f : \mathbb{R}^{n} \to \mathbb{R}$ be a scalar valued function with partitioned inputs $\uvec = [\uvec_1, \dots, \uvec_m]$ such that $\sum_{i=1}^m \dim(\uvec_i) = n$. With a slight abuse of notation we define $f(\uvec_1, \dots, \uvec_m) = f([\uvec_1, \dots, \uvec_m])$. We denote
$\nabla_{\uvec_i} f(\hat{\uvec}) \in \reals^{\dim(\uvec_i)}$ to be the $i$-th block of the gradient of $f$ evaluated at $\hat{\uvec} = [\hat{\uvec}_1, \dots, \hat{\uvec}_m]$, and further use $\frac{\diff f}{\diff \vvec}$ to denote the total derivative of $f$ with respect to $\vvec$, which exists if $\uvec$ is a differentiable function of $\vvec$. Note that in general $\nabla_{\uvec_i} f(\hat{\uvec}) \ne \frac{\diff f}{\diff \uvec_i}$ since other components of  $\uvec$ could be a function of $\uvec_i$.\footnote{This will indeed be the case in our approach: when we calculate $\ELBO(\lambda_K, \theta, \xvec)$, $\lambda_K$ is a function of the data point $\xvec$, the generative model $\theta$, and the inference network $\phi$ (Section~\ref{sec:savi}).} We also let $\Hess_{\uvec_i, \uvec_j}f(\hat{\uvec}) \in \reals^{\dim(\uvec_i) \times \dim(\uvec_j)}$ be the matrix formed by taking the $i$-th group of rows and the $j$-th group of columns of the Hessian of $f$ evaluated at $\hat{\uvec}$. 
These definitions generalize straightforwardly when $f : \mathbb{R}^n \to \mathbb{R}^p$ is a vector-valued function (e.g.  $\frac{\diff f}{\diff \uvec} \in \mathbb{R}^{n \times p}$).\footnote{Total derivatives/Jacobians are usually denoted with row vectors but we denote them with  column vectors for clearer notation.} 
\vspace{-2mm}
\subsection{Variational Inference} 
\vspace{-1mm}
Consider the following generative process for $\xvec$,
\begin{align*} 
 \zvec \sim p(\zvec) && \xvec \sim p(\xvec \given \zvec \param \theta) 
\end{align*}
where $p(\zvec)$ is the prior and $p(\xvec \given \zvec \param \theta)$ is given by a generative model with parameters $\theta$. 
As maximizing the log-likelihood $\log p(\xvec ;\theta) = \log \int_\zvec p(\xvec \given \zvec \param \theta)p(\zvec) d\zvec$ is usually intractable, variational inference instead defines a variational family of distributions $q(\zvec \param \lambda)$ parameterized by $\lambda$ and maximizes the evidence lower bound (ELBO)
\begin{align*}
\log p(\xvec ; \theta) &
\ge \E_{q(\zvec ; \lambda)} [\log p(\xvec \given \zvec)] - \KL[q(\zvec ; \lambda)\, \Vert \, p(\zvec)]\\
&= \ELBO(\lambda, \theta, \xvec)
\end{align*}
The variational posterior, $q(\zvec ; \lambda)$, is said to \textit{collapse} to the prior if $\KL[q(\zvec ; \lambda)\, \Vert \, p(\zvec)] \approx 0$.
In the general case we are given a dataset $\xvec^{(1)}, \dots, \xvec^{(N)}$ and need to find variational parameters $\lambda^{(1)}, \dots, \lambda^{(N)}$ and generative model parameters $\theta$ that jointly maximize $\sum_{i=1}^N \ELBO(\lambda^{(i)}, \theta, \xvec^{(i)})$. 
\vspace{-1mm}
\subsection{Stochastic Variational Inference}
\vspace{-1mm}
We can apply SVI \cite{Hoffman2013} with gradient ascent to approximately maximize the above objective:\footnote{While we describe the various algorithms for a specific data point, in practice we use mini-batches.}
\vspace{-2mm}
\begin{enumerate}
\item Sample $\xvec^{} \sim p_\mcD(\xvec)$   \vspace{-1mm}
\item Randomly initialize $\lambda^{}_0$ \vspace{-1mm}
\item For $k = 0, \dots, K-1$, set \\
$\lambda_{k+1}^{} = \lambda_k^{} + \alpha \nabla_{\lambda}\ELBO(\lambda_{k}^{}, \theta, \xvec^{})$ \vspace{-2mm}
\item Update $\theta$ based on $\nabla_\theta \ELBO(\lambda^{}_K , \theta, \xvec^{})$ 
\end{enumerate}
\vspace{-1mm}
Here $K$ is the number of SVI iterations and $\alpha$ is the learning rate. (Note that $\theta$ is updated based on the gradient $\nabla_\theta \ELBO(\lambda^{}_K , \theta, \xvec^{})$ and not the total derivative $\frac{\diff \ELBO(\lambda^{}_K , \theta, \xvec^{})}{\diff \theta}$. The latter would take into account the fact that $\lambda^{}_k$ is a function of $\theta$ for $k > 0$.)

SVI optimizes directly for instance-specific variational distributions, but may require running iterative inference for a large number of steps. Further, because of this block coordinate ascent approach the variational parameters $\lambda$ are optimized separately from $\theta$, 
potentially making it difficult for $\theta$ to adapt to local optima. 
\vspace{-1mm}
\subsection{Amortized Variational Inference}
\vspace{-1mm}
AVI uses a global parametric model to predict the local variational parameters for each data point. A particularly popular application of AVI is in training the variational autoencoder (VAE) \cite{Kingma2014}, which runs an inference network (i.e. encoder) $\enc(\cdot)$ parameterized by $\phi$ over the input to obtain the variational parameters:
\vspace{-2mm}
\begin{enumerate}
\item Sample $\xvec^{} \sim p_\mcD(\xvec)$ \vspace{-1mm}
\item Set $\lambda^{} = \enc(\xvec^{} \param \phi) $ \vspace{-1mm}
\item Update $\theta$ based on $\nabla_\theta \ELBO(\lambda^{} , \theta, \xvec^{})$ (which in this case is equal to  the total derivative) \vspace{-1mm}
\item Update $\phi$ based on 
\[\frac{\diff \ELBO(\lambda^{} , \theta, \xvec^{})}{\diff\phi} = \frac{\diff \lambda^{}}{\diff \phi}\nabla_{\lambda} \ELBO(\lambda^{} , \theta, \xvec^{}) \]
\end{enumerate}
\vspace{-2mm}
The inference network is learned jointly alongside the generative model with the same loss function, allowing the pair to coadapt.  
Additionally inference for AVI involves running the inference network over the input, which is usually
much faster than running iterative optimization on the ELBO. Despite these benefits, requiring the variational parameters to be a parametric function of the input may be too strict of a restriction and 
can lead to an amortization gap. This gap can propagate forward to hinder the learning of the generative model if $\theta$ is updated based on suboptimal $\lambda$.
\vspace{-3mm}
\section{Semi-Amortized Variational Autoencoders}\label{sec:savi}
\vspace{-1mm}
Semi-amortized variational autoencoders (SA-VAE) utilize an inference network over the input to give the initial variational parameters, and subsequently run SVI to refine them. One might appeal to the universal approximation theorem \cite{Hornik1989} and question the necessity of additional SVI steps given a rich-enough inference network. However, in practice we find that the variational parameters found from VAE are usually not optimal even with a powerful inference network, and the amortization gap can be significant especially on complex datasets \cite{Cremer2017,Krishnan2017}. 

SA-VAE models are trained using a combination of AVI and SVI steps:
\vspace{-2mm}
\begin{enumerate}
\item Sample $\xvec^{} \sim p_\mcD(\xvec)$ \vspace{-1mm}
\item Set $\lambda^{}_0 = \enc(\xvec^{} \param \phi) $  \vspace{-1mm}
\item For $k = 0, \dots, K-1$, set \\
$\lambda_{k+1}^{} = \lambda_k^{} + \alpha \nabla_{\lambda}\ELBO(\lambda_{k}^{}, \theta, \xvec^{})$ \vspace{-1mm}
\item Update $\theta$ based on $\frac{\diff \ELBO(\lambda_K^{}, \theta, \xvec^{})}{\diff \theta}$ \vspace{-1mm}
\item Update $\phi$ based on $\frac{\diff \ELBO(\lambda_K^{}, \theta, \xvec^{})}{\diff \phi}$
\end{enumerate}
\vspace{-2mm}
Note that for training we need to compute the total derivative of the final ELBO with respect to $\theta, \phi$ (i.e. steps 4 and 5 above).
Unlike with AVI, in order to update the encoder and generative model parameters, this total derivative requires backpropagating through 
the SVI updates. Specifically this requires backpropagating through gradient ascent \cite{Domke2012,Maclaurin2015}. 

Following past work, this backpropagation step can be done efficiently with fast Hessian-vector products \cite{LeCun1993,Pearlmutter1994}. 
In particular, consider the case where we perform one step of refinement,
$\lambda_1 = \lambda_0 + \alpha \nabla_{\lambda}\ELBO(\lambda_0, \theta, \xvec)$, and for brevity let $\mcL = \ELBO(\lambda_1, \theta, \xvec)$. To backpropagate through this, we receive the derivative $\frac{\diff \mcL}{\diff \lambda_1}$ and use the chain rule,
\begin{align*}
\frac{\diff \mcL}{\diff \lambda_0} &= \frac{\diff \lambda_1}{\diff \lambda_0}\frac{\diff \mcL}{\diff \lambda_1}  = (\mathbf{I} + \alpha\Hess_{\lambda,\lambda}\ELBO(\lambda_0, \theta, \xvec))\frac{\diff \mcL}{\diff \lambda_1} \\
&= \frac{\diff \mcL}{\diff \lambda_1} + \alpha\Hess_{\lambda,\lambda}\ELBO(\lambda_0, \theta, \xvec)\frac{\diff \mcL}{\diff \lambda_1}
\end{align*}
We can then backpropagate $\frac{\diff \mcL}{\diff \lambda_0}$ through the inference network to calculate the total derivative, i.e.
$\frac{\diff \mcL}{\diff \phi} = \frac{\diff \lambda_0}{\diff \phi}\frac{\diff \mcL}{\diff \lambda_0}$. Similar rules can be used to derive $\frac{\diff \mcL}{\diff \theta}$.\footnote{We refer the reader to \citet{Domke2012} for the full derivation.} The full forward/backward step, which uses gradient descent with momentum on the negative ELBO, is shown in Algorithm~\ref{alg:savi}.

\begin{algorithm}[tb]

   \caption{Semi-Amortized Variational Autoencoders}
   \label{alg:savi}
\begin{algorithmic}
   \STATE {\bfseries Input:} inference network $\phi$, generative model $\theta$, \\ 
   \hspace{10mm} inference steps $K$, learning rate $\alpha$, momentum $\gamma$,   \\
   \hspace{10mm} loss function $f(\lambda, \theta, \xvec) = -\ELBO(\lambda, \theta, \xvec)$

   \STATE Sample $\xvec^{} \sim p_\mathcal{D}(\xvec)$
   \STATE $\lambda_0 \gets \enc(\xvec^{} \param \phi)$ 
   \STATE $v_0 \gets 0$
   \FOR{$k=0$ {\bfseries to} $K-1$} 
   \STATE $v_{k+1} \gets \gamma v_{k} - \nabla_{\lambda}f(\lambda_k, \theta, \xvec^{})$
   \STATE $\lambda_{k+1} \gets \lambda_k + \alpha v_{k+1}$
   \ENDFOR
   \STATE  $\mcL \gets f(\lambda_K, \theta, \xvec^{})$ 
   \STATE $\overline{\lambda}_K \gets \nabla_{\lambda}f(\lambda_K, \theta, \xvec^{}) $
   \STATE $\overline{\theta} \gets \nabla_{\theta}f(\lambda_K, \theta, \xvec^{})$
   \STATE $\overline{v}_K \gets 0$
   \FOR{$k=K-1$ {\bfseries to} $0$}
   \STATE $\overline{v}_{k+1} \gets \overline{v}_{k+1} + \alpha \overline{\lambda}_{k+1}$
   \STATE $\overline{\lambda}_k \gets \overline{\lambda}_{k+1} - \Hess_{\lambda, \lambda}f(\lambda_k, \theta, \xvec^{}) \overline{v}_{k+1} $
   \STATE $\overline{\theta} \gets \overline{\theta} - \Hess_{\theta,\lambda}f(\lambda_k, \theta, \xvec^{}) \overline{v}_{k+1}$
   \STATE $\overline{v}_{k} \gets \gamma \overline{v}_{k+1}$
   \ENDFOR
   \STATE $\frac{\diff \mcL}{\diff \theta} \gets \overline{\theta} $   
   \STATE $\frac{\diff \mcL}{\diff \phi} \gets \frac{\diff \lambda_0}{\diff \phi}\overline{\lambda}_0 $ 
   \STATE Update $\theta, \phi$ based on $\frac{\diff \mcL}{\diff \theta}, \frac{\diff \mcL}{\diff \phi}$
\end{algorithmic}
\end{algorithm}

In our implementation we calculate Hessian-vector products with finite differences \cite{LeCun1993,Domke2012}, which was found to be more memory-efficient than automatic differentiation (and therefore crucial for scaling our approach to rich inference networks/generative models). Specifically, we estimate $\Hess_{\uvec_i, \uvec_j}f(\hat{\uvec}) \vvec$ with
\begin{align*}
\Hess_{\uvec_i, \uvec_j}f(\hat{\uvec}) \vvec \approx  \frac{1}{\epsilon}\Big(& \nabla_{\uvec_i} f(\hat{\uvec}_0, \dots, \hat{\uvec}_j + \epsilon \vvec, \dots,\hat{\uvec}_m) \\ 
&- \nabla_{\uvec_i}f(\hat{\uvec}_0, \dots, \hat{\uvec}_j \dots,\hat{\uvec}_m) \Big)
\end{align*}
where $\epsilon$ is some small number (we use $\epsilon = 10^{-5}$).\footnote{Since in our case the ELBO is a non-deterministic function due to sampling (and dropout, if applicable), care must be taken when calculating Hessian-vector product with finite differences to ensure that the source of randomness is the same when calculating the two gradient expressions.} We further clip the results (i.e. rescale the results if the norm exceeds a threshold) before and after each Hessian-vector product as well as during SVI, which helped mitigate exploding gradients and further gave better training signal to the inference network.\footnote{Without gradient clipping, in addition to numerical issues we empirically observed the model to degenerate to a case whereby it learned to rely too much on iterative inference, and thus the initial parameters from the inference network were poor. Another way to provide better signal to the inference network is to train against a weighted sum 
$\sum_{k=0}^K w_k \ELBO (\lambda_k, \theta, \xvec)$ for $w_k \ge 0$.} See Appendix A for details. 

\section{Experiments}
\vspace{-2mm}
We apply our approach to train generative models on a synthetic dataset in addition to text/images. For all experiments we utilize stochastic gradient descent with momentum on the negative ELBO. Our prior is the spherical Gaussian $\mcN(\mathbf{0}, \mathbf{I})$ and the variational posterior is diagonal Gaussian, where the variational parameters are given by the mean vector and the diagonal log variance vector, i.e. $\lambda = [\boldsymbol{\mu}, \log \boldsymbol{\sigma}^2]$. 

In preliminary experiments we also experimented with natural gradients, other optimization algorithms, and learning the learning rates, but found that these did not significantly improve results. Full details regarding hyperparameters/model architectures for all experiments are in Appendix B.
\vspace{-3mm}
\subsection{Synthetic Data}
\vspace{-1mm}
\begin{figure}
\centering
\vspace{-3mm}
\includegraphics[scale=0.38]{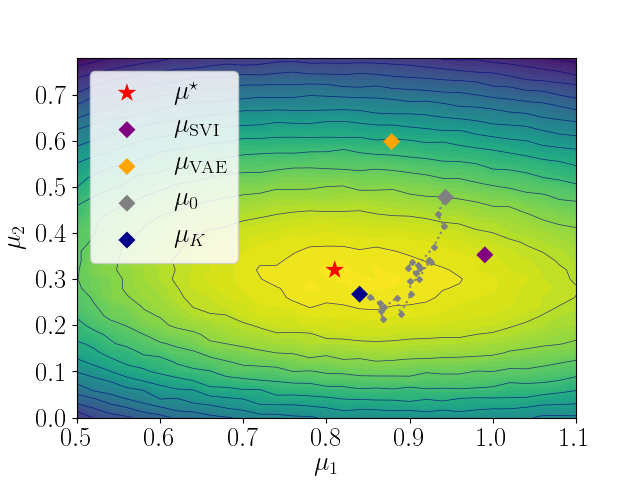} 
\vspace{-4.5mm}
\caption{ELBO landscape with the oracle generative model as a function of the variational posterior means $\mu_1, \mu_2$ for a randomly chosen test point. Variational parameters obtained from VAE, SVI are shown as $\mu_{\text{VAE}}, \mu_{\text{SVI}}$ and the initial/final parameters from SA-VAE are shown as $\mu_{0}$ and $\mu_{K}$ (along with the intermediate points). SVI/SA-VAE are run for 20 iterations. The optimal point, found from grid search, is shown as $\mu^\star$.}
\label{fig:toy}
\vspace{-4mm}
\end{figure}
We first apply our approach to a synthetic dataset where we have access to the true underlying generative model of discrete sequences. We generate synthetic sequential data according to the following oracle generative process with 2-dimensional latent variables and $\xvec_t$:
\begin{align*}
z_1, z_2 &\sim \mcN (0, 1) \,\,\,\,\,\,\,\,\,\,\,\,\,\, \hvec_t = \LSTM(\hvec_{t-1}, \xvec_t)  \\
&\xvec_{t+1} \sim \softmax(\MLP([\hvec_t , z_1, z_2]))
\end{align*}
We initialize the LSTM/MLP randomly as $\theta$, where the LSTM has a single layer with hidden state/input dimension equal to 100. We generate for 5 time steps (so each example is given by $\xvec = [\xvec_1, \dots, \xvec_5]$) with a vocabulary size of 1000 for each $\xvec_t$. Training set consists of 5000 points. See Appendix B.1 for the exact setup.

We fix this oracle generative model $p(\xvec \given \zvec \param \theta)$ and learn an inference network (also a one-layer LSTM) with VAE and SA-VAE.\footnote{With a fixed oracle, these models are technically not VAEs as VAE usually implies that the  the generative model is learned (alongside the encoder).} For a randomly selected test point, we plot the ELBO landscape in Figure~\ref{fig:toy} as a function of the variational posterior means ($\mu_1, \mu_2$) learned from the different methods. For SVI/SA-VAE we run iterative optimization for 20 steps.
Finally we also show the optimal variational parameters found from grid search. 

\begin{table}
\centering
\begin{sc}
\begin{small}
\vspace{-1mm}
\begin{tabular}{l r r}
\toprule
Model & Oracle Gen & Learned Gen  \\
\midrule
VAE & $ \le 21.77$  & $\le 27.06$  \\
SVI & $ \le 22.33 $ &  $\le 25.82$ \\
SA-VAE & $ \le  20.13 $ & $\le 25.21$ \\
\midrule
True NLL (Est)  & $ 19.63 $ & $-$  \\
\bottomrule
\end{tabular}
\end{small}
\end{sc}
\vspace{-2mm}
\caption{Variational upper bounds for the various models on the synthetic dataset, where SVI/SA-VAE is trained/tested with 20 steps. \textsc{True NLL (Est)} is an estimate of the true negative log-likelihood (i.e. entropy of the data-generating distribution) estimated with 1000 samples from the prior. $\textsc{Oracle Gen}$ uses the oracle generative model and $\textsc{Learned Gen}$ learns the generative network.}
\label{tab:toy}
\vspace{-6mm}
\end{table}

As can be seen from Figure~\ref{fig:toy}, the variational parameters from running SA-VAE are closest to the optimum while those obtained from SVI and VAE are slightly further away.
In Table~\ref{tab:toy} we show the variational upper bounds (i.e. negative ELBO) on the negative log-likelihood (NLL) from training the various models with
both the oracle/learned generative model, and find that SA-VAE outperforms VAE/SVI in both cases.
\vspace{-2mm}
\subsection{Text}
\vspace{-2mm}
The next set of experiments is focused on text modeling on the Yahoo questions corpus from \citet{Yang2017}. Text modeling with deep generative models has been a challenging problem, and few approaches have been shown to produce rich generative models that do not collapse to standard language models. Ideally a deep generative model trained with variational inference would make use of the latent space (i.e. maintain a nonzero KL term) while accurately modeling the underlying distribution. 

Our architecture and hyperparameters are identical to the LSTM-VAE baselines considered in \citet{Yang2017}, except that we train with SGD instead of Adam, which was found to perform better for training LSTMs. Specifically, both the inference network and the generative model are one-layer LSTMs with 1024 hidden units and 512-dimensional word embeddings. The last hidden state of the encoder is used to predict the vector of variational posterior means/log variances. The sample from the variational posterior is used to predict the initial hidden state of the generative LSTM and additionally fed as input at each time step. The latent variable is 32-dimensional.  Following previous works \cite{Bowman2016,Son2016,Yang2017}, for all the variational models we utilize a KL-cost annealing strategy whereby the multiplier on the KL term is increased linearly from 0.1 to 1.0 each batch over 10 epochs. Appendix B.2 has the full architecture/hyperparameters.

In addition to autoregressive/VAE/SVI baselines, we consider two other approaches that also combine amortized inference with iterative refinement. 
The first approach is from \citet{Krishnan2017}, where the generative model takes a gradient step based on the final variational parameters and the inference network takes a gradient step based on the initial variational parameters, i.e. we update $\theta$ based on $\nabla_\theta\ELBO(\lambda_K^{}, \theta, \xvec^{})$ and update $\phi$ based on $ \frac{\diff \lambda_0^{}}{\diff \phi}\nabla_\lambda\ELBO(\lambda_0^{}, \theta, \xvec^{})$.
The forward step (steps 1-3 in Section~\ref{sec:savi}) is identical to SA-VAE.
We refer to this baseline as VAE + SVI.  

In the second approach, based on \citet{Salakhutdinov2010} and \citet{Hjelm2016}, we train the inference network to minimize the KL-divergence between the initial and the final variational distributions, keeping the latter fixed. Specifically, letting $g(\nu, \omega) =\KL[q(\zvec \param \nu ) \, \Vert \, q(\zvec \param \omega)]$, we update $\theta$ based on $\nabla_\theta\ELBO(\lambda_K^{}, \theta, \xvec^{})$ and update $\phi$ based on $\frac{\diff \lambda_0^{}}{\diff \phi}\nabla_\nu g(\lambda_0^{}, \lambda_K^{})$.
Note that the inference network is not updated based on
$\frac{\diff g}{\diff \phi}$, which would take into account the fact that both $\lambda_0$ and $\lambda_K$ are functions of $\phi$. We found $g(\lambda_0, \lambda_K)$ to perform better than the reverse direction $g(\lambda_K, \lambda_0)$. We refer to this setup as VAE + SVI + KL.

\begin{table}
\centering
\begin{sc}
\begin{small}
\vspace{-1mm}
\begin{tabular}{l r r r r}
\toprule
Model &  NLL & KL & PPL \\
\midrule
LSTM-LM & $334.9$ & $-$ & $66.2$ \\
LSTM-VAE  & $\le 342.1 $ & $0.0$ & $\le 72.5$ \\
LSTM-VAE + Init & $\le 339.2$ & $0.0$ & $\le 69.9$\\
CNN-LM & $335.4$ & $-$ & $66.6$ \\
CNN-VAE & $\le 333.9 $ & $6.7$ & $\le 65.4$ \\
CNN-VAE + Init & $\le 332.1 $ & $10.0$ & $\le 63.9$ \\
\midrule
LM & $ 329.1 $ & $-$ & $ 61.6$ \\
VAE & $ \le 330.2 $ & $0.01$ & $\le 62.5$ \\
VAE + Init & $\le 330.5 $ & $0.37 $ & $\le 62.7$ \\
VAE + Word-Drop 25\% & $\le 334.2 $ & $1.44 $ & $\le 65.6$ \\
VAE + Word-Drop 50\% & $\le 345.0 $ & $5.29 $ & $\le 75.2$ \\
SVI ($K=10$) & $\le 331.4$ & $0.16$ & $\le 63.4$ \\
SVI ($K=20$) & $\le 330.8 $ & $0.41$ & $\le 62.9 $ \\
SVI ($K=40$) & $\le 329.8 $ & $1.01$ & $\le 62.2 $ \\
VAE + SVI ($K=10$) & $ \le 331.2$ & $7.85$ & $\le 63.3 $ \\
VAE + SVI ($K=20$) & $\le 330.5$ & $7.80$ & $\le 62.7$ \\
VAE + SVI + KL ($K=10$) & $\le 330.3 $ & $7.95$ & $\le 62.5$ \\
VAE + SVI + KL ($K=20$) & $\le 330.1$ & $7.81$ & $\le 62.3$ \\
SA-VAE ($K=10$) &  $\le 327.6$ & $5.13$ & $\le 60.5$ \\
SA-VAE ($K=20$) &  $\le 327.5$ & $7.19$ & $\le 60.4$ \\
\bottomrule
\vspace{-5mm}
\end{tabular}
\end{small}
\end{sc}
\caption{Results on text modeling on the Yahoo dataset. Top results are from \citet{Yang2017}, while the bottom results are from this work (\textsc{+ Init} means the encoder is initialized with a pretrained language model, while models with \textsc{+ Word-Drop} are trained with word-dropout). NLL/KL numbers are averaged across examples, and PPL refers to perplexity. $K$ refers to the number of inference steps used for training/testing.}
\label{tab:yahoo}
\vspace{-6mm}
\end{table}

Results from the various models are shown in Table~\ref{tab:yahoo}.
Our baseline models (LM/VAE/SVI in Table~\ref{tab:yahoo}) are already quite strong and outperform the models considered in \citet{Yang2017}. However models trained with VAE/SVI make negligible use of the latent variable and practically collapse to a language model, negating the benefits of using latent variables.\footnote{Models trained with word dropout (+ \textsc{Word-Drop} in Table~\ref{tab:yahoo}) do make use of the latent space but significantly underperform a language model.} In contrast, models that combine amortized inference with iterative refinement make use of the latent space and the KL term is significantly above zero.\footnote{A high KL term does not necessarily imply that the latent variable is being utilized in a meaningful way (it could simply be due to bad optimization). In Section~\ref{sec:latent} we investigate the learned latent space in more detail.}
VAE + SVI and VAE + SVI + KL do not outperform a language model, and while SA-VAE only modestly outperforms it, to our knowledge this is one of the first instances in which we are able to train an LSTM generative model that does not ignore the latent code and outperforms a language model.

\begin{figure}
\centering
\begin{subfigure}{0.23\textwidth}
\centering
\includegraphics[scale=0.27]{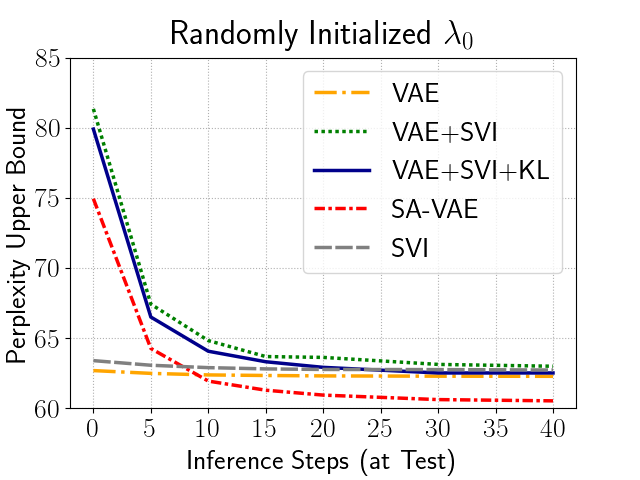} 
\end{subfigure}
\begin{subfigure}{0.23\textwidth}
\hspace{-3mm}
\centering
\includegraphics[scale=0.27]{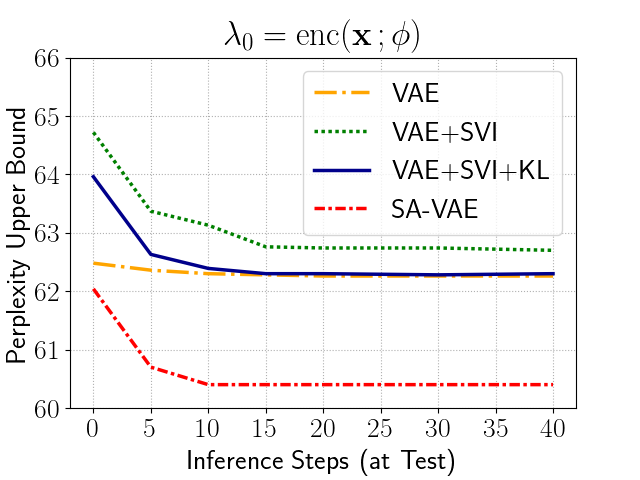} 
\end{subfigure}
\vspace{-2mm}
\caption{(Left) Perplexity upper bound of various models when trained with 20 steps (except for VAE) and tested with varying number of SVI steps from random initialization. (Right) Same as the left except that SVI is initialized with variational parameters obtained from the inference network.}
\label{fig:ppl}
\vspace{-6mm}
\end{figure}

One might wonder if the improvements are coming from simply having a more flexible inference scheme at test time, rather than from learning a better generative model. To test this, for the various models we discard the inference network at test time and perform SVI for a variable number of steps from random initialization. The results are shown in Figure~\ref{fig:ppl} (left). It is clear that the learned generative model (and the associated ELBO landscape) is quite different---it is not possible to train with VAE and perform SVI at test time to obtain the same performance as SA-VAE (although the performance of VAE does improve slightly from 62.7 to 62.3 when we run SVI for 40 steps from random initialization). 

Figure~\ref{fig:ppl} (right) has the results for a similar experiment where we refine the variational parameters initialized from the inference network for a variable number of steps at test time. We find that the inference network provides better initial parameters than random initialization and thus requires fewer iterations of SVI to reach the optimum. We do not observe improvements for running more refinement steps than was used in training at test time. Interestingly, SA-VAE without any refinement steps at test time has a substantially nonzero KL term (KL = 6.65, PPL = 62.0). This indicates that the posterior-collapse phenomenon when training LSTM-based VAEs for text is partially due to optimization issues.
Finally, while \citet{Yang2017} found that initializing the encoder with a pretrained language model improved performance (\textsc{+ Init} in Table~\ref{tab:yahoo}), we did not observe this on our baseline VAE model when we trained with SGD and hence did not pursue this further.

\begin{table}
\center
\begin{sc}
\begin{small}
\begin{tabular}{l r}
\toprule
Model & NLL \\
\midrule
IWAE \textnormal{\cite{Burda2015}} & $103.38$ \\
Ladder VAE \textnormal{\cite{Son2016}} & $102.11$ \\
RBM \textnormal{\cite{Burda2015b}} & $100.46$ \\
Discrete VAE \textnormal{\cite{Rolfe2017}} & $97.43$ \\
DRAW \textnormal{\cite{Gregor2015}} & $\le 96.50$ \\
Conv DRAW \textnormal{\cite{Gregor2016}} & $\le 91.00$ \\
VLAE \textnormal{\cite{Chen2017}} & $89.83$ \\
VampPrior \textnormal{\cite{Tomczak2017}} & $89.76$ \\
\bottomrule
Gated PixelCNN & $90.59$ \\
VAE & $\le 90.43 \,\,(0.98)$ \\
SVI ($K=10$) & $\le 90.65 \,\,(0.02)$\\
SVI ($K=20$) & $\le 90.51 \,\,(0.06)$  \\
SVI ($K=40$) & $\le 90.44 \,\,(0.27)$  \\
SVI ($K=80$) & $\le 90.27 \,\,(1.65)$  \\
VAE + SVI ($K=10$) & $ \le 90.26 \,\,(1.69) $ \\
VAE + SVI ($K=20$) & $\le 90.19 \,\,(2.40) $ \\
VAE + SVI + KL ($K=10$) & $\le 90.24 \,\,(2.42) $  \\
VAE + SVI + KL ($K=20$) & $\le 90.21 \,\,(2.83) $ \\
SA-VAE ($K=10$) &  $\le 90.20 \,\,(1.83)$ \\
SA-VAE ($K=20$) &  $\le 90.05 \,\,(2.78)$ \\
\bottomrule
\vspace{-7mm}
\end{tabular}
\end{small}
\end{sc}
\caption{Results on image modeling on the OMNIGLOT dataset. Top results are from prior works, while the bottom results are from this work. \textsc{Gated PixelCNN} is our autoregressive baseline, and $K$ refers to the number of inference steps during training/testing. For the variational models the KL portion of the ELBO is shown in parentheses.}
\vspace{-7mm}
\label{tab:omni}
\end{table}

\vspace{-3mm}
\subsection{Images}
\vspace{-2mm}
We next apply our approach to model images on the OMNIGLOT dataset \cite{Lake2015}.\footnote{We focus on the more complex OMNIGLOT dataset instead of the simpler MNIST dataset as prior work has shown that the amortization gap on MNIST is minimal \cite{Cremer2017}.} While posterior collapse is less of an issue for VAEs trained on images, we still expect that improving the amortization gap would result in generative models that better model the underlying data and make more use of the latent space. We use a three-layer ResNet \cite{He2016} as our inference network. The generative model first transforms the 32-dimensional latent vector to the image spatial resolution, which is concatenated with the original image and fed to a 12-layer Gated PixelCNN \cite{Oord2016b} with varying filter sizes, followed by a final sigmoid layer. We employ the same KL-cost annealing schedule as in the text experiments. See Appendix B.3 for the exact architecture/hyperparameters.

Results from the various models are shown in Table~\ref{tab:omni}. Our findings are largely consistent with results from text: the semi-amortized approaches outperform VAE/SVI baselines, and further they learn generative models that make more use of the latent representations (i.e. KL portion of the loss is higher). Even with 80 steps of SVI we are unable to perform as well as SA-VAE trained with 10 refinement steps, indicating the importance of good initial parameters provided by the inference network. In Appendix C we further investigate the performance of VAE and SA-VAE as we vary the training set size and the capacity of the inference network/generative model. We find that SA-VAE outperforms VAE and has higher latent variable usage in all scenarios.
We note that we do not outperform the state-of-the-art models that employ hierarchical latent variables and/or more sophisticated priors \cite{Chen2017,Tomczak2017}. However these additions are largely orthogonal to our approach and we hypothesize they will also benefit from combining amortized inference with iterative refinement.\footnote{Indeed, \citet{Cremer2017} observe that the amortization gap can be substantial for VAE trained with richer variational families.}

\vspace{-3mm}
\section{Discussion}
\vspace{-2mm}
\subsection{Learned Latent Space}\label{sec:latent}
\vspace{-2mm}
For the text model we investigate what the latent variables are learning through saliency analysis with our best model (SA-VAE trained with 20 steps). Specifically, we calculate the output saliency of each token $\xvec_t$ with respect to $\zvec$ as
\begin{align*}
\vspace{-1mm}
\E_{q(\zvec^{} \param \lambda^{})}\Big[ \, \Big\Vert \, \frac{\diff \log p(\xvec^{}_t \given \xvec^{}_{< t}, \zvec^{} \param \theta)}{\diff \zvec}\, \Big\Vert_2\, \Big]
\end{align*}
where $\Vert \cdot \Vert_2$ is the $l_2$ norm and the expectation is approximated with  5 samples from the variational posterior. Saliency is therefore a measure of how much the latent variable is being used to predict a particular token.

\begin{figure*}[t]
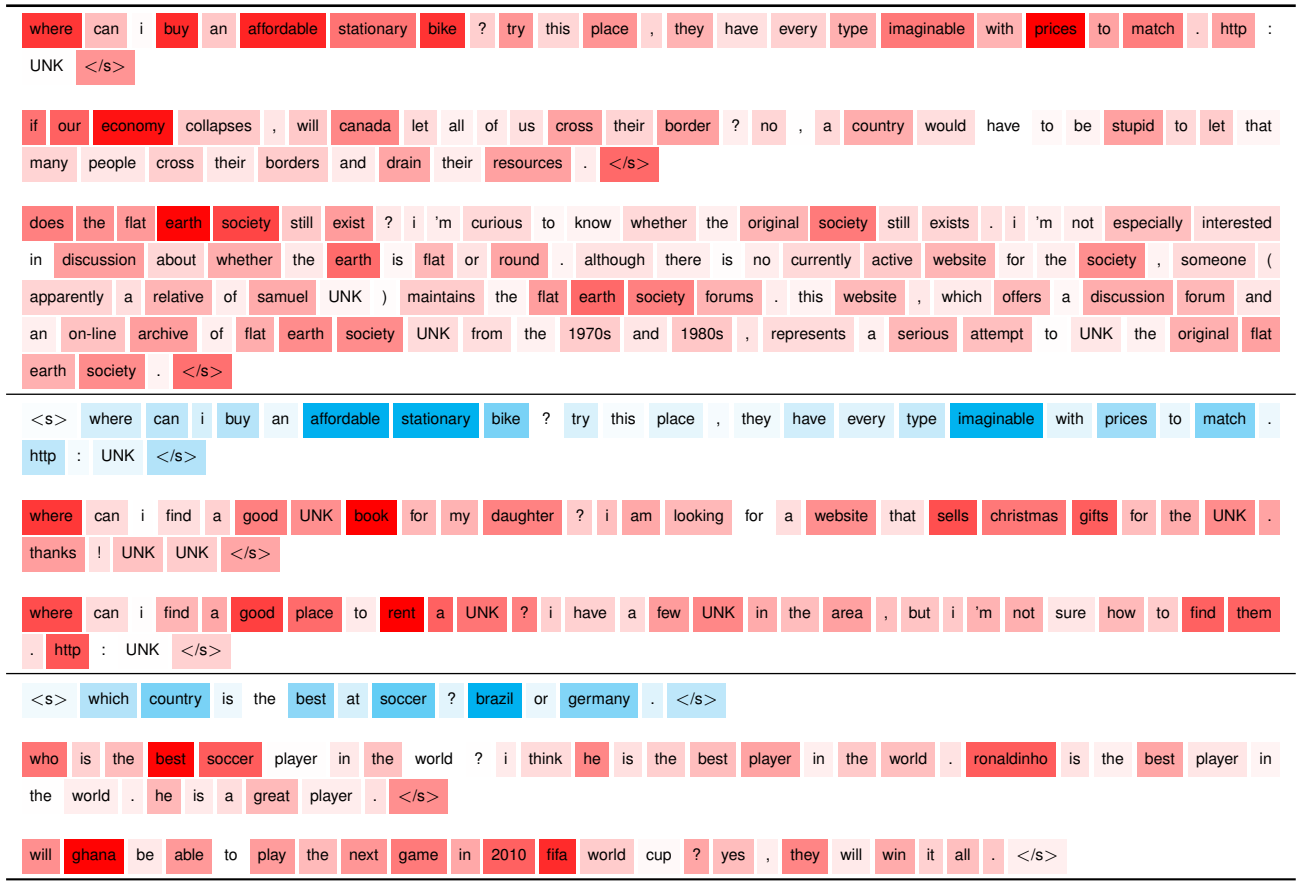

\vspace{-2.5mm}
\begin{tabularx}{\textwidth}{X}
\textsf{\tiny 
\colorbox{red!79}{\strut where} \colorbox{red!26}{\strut can} \colorbox{red!1.6}{\strut i} \colorbox{red!83}{\strut buy} \colorbox{red!22}{\strut an} \colorbox{red!86}{\strut affordable} \colorbox{red!55}{\strut stationary} \colorbox{red!80}{\strut bike} \colorbox{red!19}{\strut ?} \colorbox{red!41}{\strut try} \colorbox{red!16}{\strut this} \colorbox{red!42}{\strut place} \colorbox{red!16}{\strut ,} \colorbox{red!30}{\strut they} \colorbox{red!12}{\strut have} \colorbox{red!18}{\strut every} \colorbox{red!31}{\strut type} \colorbox{red!53}{\strut imaginable} \colorbox{red!30}{\strut with} \colorbox{red!100}{\strut prices} \colorbox{red!36}{\strut to} \colorbox{red!51}{\strut match} \colorbox{red!18}{\strut .} \colorbox{red!36}{\strut http} \colorbox{red!0}{\strut :} \colorbox{red!0.22}{\strut UNK} \colorbox{red!42}{\strut $<$/s$>$}
}
\\ 
\vspace{0.25mm}
\textsf{\tiny 
\colorbox{red!55}{\strut if} \colorbox{red!62}{\strut our} \colorbox{red!93}{\strut economy} \colorbox{red!21}{\strut collapses} \colorbox{red!9.7}{\strut ,} \colorbox{red!21}{\strut will} \colorbox{red!47}{\strut canada} \colorbox{red!14}{\strut let} \colorbox{red!5.6}{\strut all} \colorbox{red!8.1}{\strut of} \colorbox{red!12}{\strut us} \colorbox{red!36}{\strut cross} \colorbox{red!17}{\strut their} \colorbox{red!37}{\strut border} \colorbox{red!6.6}{\strut ?} \colorbox{red!16}{\strut no} \colorbox{red!0.42}{\strut ,} \colorbox{red!15}{\strut a} \colorbox{red!32}{\strut country} \colorbox{red!13}{\strut would} \colorbox{red!0}{\strut have} \colorbox{red!2.8}{\strut to} \colorbox{red!4.9}{\strut be} \colorbox{red!37}{\strut stupid} \colorbox{red!15}{\strut to} \colorbox{red!20}{\strut let} \colorbox{red!4}{\strut that} \colorbox{red!17}{\strut many} \colorbox{red!10}{\strut people} \colorbox{red!19}{\strut cross} \colorbox{red!9.8}{\strut their} \colorbox{red!20}{\strut borders} \colorbox{red!8.4}{\strut and} \colorbox{red!35}{\strut drain} \colorbox{red!7.2}{\strut their} \colorbox{red!37}{\strut resources} \colorbox{red!7.5}{\strut .} \colorbox{red!59}{\strut $<$/s$>$}
}
\\ 

\vspace{0.25mm}
\textsf{\tiny
\colorbox{red!50}{\strut does} \colorbox{red!43}{\strut the} \colorbox{red!40}{\strut flat} \colorbox{red!98}{\strut earth} \colorbox{red!73}{\strut society} \colorbox{red!21}{\strut still} \colorbox{red!39}{\strut exist} \colorbox{red!9.2}{\strut ?} \colorbox{red!7.8}{\strut i} \colorbox{red!7}{\strut 'm} \colorbox{red!15}{\strut curious} \colorbox{red!3.3}{\strut to} \colorbox{red!4.9}{\strut know} \colorbox{red!15}{\strut whether} \colorbox{red!6.1}{\strut the} \colorbox{red!29}{\strut original} \colorbox{red!52}{\strut society} \colorbox{red!14}{\strut still} \colorbox{red!16}{\strut exists} \colorbox{red!12}{\strut .} \colorbox{red!11}{\strut i} \colorbox{red!6.1}{\strut 'm} \colorbox{red!7.2}{\strut not} \colorbox{red!25}{\strut especially} \colorbox{red!13}{\strut interested} \colorbox{red!0}{\strut in} \colorbox{red!38}{\strut discussion} \colorbox{red!16}{\strut about} \colorbox{red!32}{\strut whether} \colorbox{red!8.8}{\strut the} \colorbox{red!58}{\strut earth} \colorbox{red!3.2}{\strut is} \colorbox{red!28}{\strut flat} \colorbox{red!11}{\strut or} \colorbox{red!35}{\strut round} \colorbox{red!4.5}{\strut .} \colorbox{red!11}{\strut although} \colorbox{red!6.7}{\strut there} \colorbox{red!1.2}{\strut is} \colorbox{red!7.2}{\strut no} \colorbox{red!15}{\strut currently} \colorbox{red!21}{\strut active} \colorbox{red!29}{\strut website} \colorbox{red!5.3}{\strut for} \colorbox{red!6.3}{\strut the} \colorbox{red!44}{\strut society} \colorbox{red!4.6}{\strut ,} \colorbox{red!15}{\strut someone} \colorbox{red!8.9}{\strut (} \colorbox{red!17}{\strut apparently} \colorbox{red!13}{\strut a} \colorbox{red!29}{\strut relative} \colorbox{red!7}{\strut of} \colorbox{red!33}{\strut samuel} \colorbox{red!3.9}{\strut UNK} \colorbox{red!1.1}{\strut )} \colorbox{red!20}{\strut maintains} \colorbox{red!4.2}{\strut the} \colorbox{red!38}{\strut flat} \colorbox{red!59}{\strut earth} \colorbox{red!50}{\strut society} \colorbox{red!30}{\strut forums} \colorbox{red!7.1}{\strut .} \colorbox{red!4.7}{\strut this} \colorbox{red!27}{\strut website} \colorbox{red!9.1}{\strut ,} \colorbox{red!7.8}{\strut which} \colorbox{red!27}{\strut offers} \colorbox{red!3}{\strut a} \colorbox{red!32}{\strut discussion} \colorbox{red!28}{\strut forum} \colorbox{red!9.9}{\strut and} \colorbox{red!6.4}{\strut an} \colorbox{red!24}{\strut on-line} \colorbox{red!38}{\strut archive} \colorbox{red!6.2}{\strut of} \colorbox{red!32}{\strut flat} \colorbox{red!45}{\strut earth} \colorbox{red!45}{\strut society} \colorbox{red!14}{\strut UNK} \colorbox{red!13}{\strut from} \colorbox{red!5.5}{\strut the} \colorbox{red!20}{\strut 1970s} \colorbox{red!10}{\strut and} \colorbox{red!22}{\strut 1980s} \colorbox{red!11}{\strut ,} \colorbox{red!14}{\strut represents} \colorbox{red!5.7}{\strut a} \colorbox{red!28}{\strut serious} \colorbox{red!25}{\strut attempt} \colorbox{red!0.93}{\strut to} \colorbox{red!9.6}{\strut UNK} \colorbox{red!5.5}{\strut the} \colorbox{red!31}{\strut original} \colorbox{red!37}{\strut flat} \colorbox{red!31}{\strut earth} \colorbox{red!45}{\strut society} \colorbox{red!4.7}{\strut .} \colorbox{red!56}{\strut $<$/s$>$}
} \\
\midrule
\textsf{\tiny 
\colorbox{cyan!1.7}{\strut $<$s$>$} \colorbox{cyan!15}{\strut where} \colorbox{cyan!23}{\strut can} \colorbox{cyan!14}{\strut i} \colorbox{cyan!26}{\strut buy} \colorbox{cyan!6.1}{\strut an} \colorbox{cyan!96}{\strut affordable} \colorbox{cyan!93}{\strut stationary} \colorbox{cyan!46}{\strut bike} \colorbox{cyan!0}{\strut ?} \colorbox{cyan!13}{\strut try} \colorbox{cyan!2.7}{\strut this} \colorbox{cyan!8.5}{\strut place} \colorbox{cyan!3.3}{\strut ,} \colorbox{cyan!4.9}{\strut they} \colorbox{cyan!16}{\strut have} \colorbox{cyan!11}{\strut every} \colorbox{cyan!20}{\strut type} \colorbox{cyan!91}{\strut imaginable} \colorbox{cyan!9.5}{\strut with} \colorbox{cyan!34}{\strut prices} \colorbox{cyan!7.1}{\strut to} \colorbox{cyan!43}{\strut match} \colorbox{cyan!6.3}{\strut .} \colorbox{cyan!23}{\strut http} \colorbox{cyan!3.7}{\strut :} \colorbox{cyan!6.9}{\strut UNK} \colorbox{cyan!27}{\strut $<$/s$>$}
} \\
\vspace{0.25mm}
\textsf{\tiny 
\colorbox{red!78}{\strut where} \colorbox{red!14}{\strut can} \colorbox{red!0.23}{\strut i} \colorbox{red!13}{\strut find} \colorbox{red!18}{\strut a} \colorbox{red!50}{\strut good} \colorbox{red!44}{\strut UNK} \colorbox{red!100}{\strut book} \colorbox{red!28}{\strut for} \colorbox{red!30}{\strut my} \colorbox{red!49}{\strut daughter} \colorbox{red!20}{\strut ?} \colorbox{red!25}{\strut i} \colorbox{red!34}{\strut am} \colorbox{red!28}{\strut looking} \colorbox{red!0}{\strut for} \colorbox{red!13}{\strut a} \colorbox{red!44}{\strut website} \colorbox{red!12}{\strut that} \colorbox{red!69}{\strut sells} \colorbox{red!53}{\strut christmas} \colorbox{red!63}{\strut gifts} \colorbox{red!26}{\strut for} \colorbox{red!32}{\strut the} \colorbox{red!45}{\strut UNK} \colorbox{red!36}{\strut .} \colorbox{red!38}{\strut thanks} \colorbox{red!12}{\strut !} \colorbox{red!24}{\strut UNK} \colorbox{red!20}{\strut UNK} \colorbox{red!35}{\strut $<$/s$>$}
} \\
\vspace{0.25mm}
\textsf{\tiny 
\colorbox{red!70}{\strut where} \colorbox{red!14}{\strut can} \colorbox{red!0.6}{\strut i} \colorbox{red!29}{\strut find} \colorbox{red!34}{\strut a} \colorbox{red!75}{\strut good} \colorbox{red!58}{\strut place} \colorbox{red!8.4}{\strut to} \colorbox{red!100}{\strut rent} \colorbox{red!62}{\strut a} \colorbox{red!54}{\strut UNK} \colorbox{red!50}{\strut ?} \colorbox{red!19}{\strut i} \colorbox{red!18}{\strut have} \colorbox{red!15}{\strut a} \colorbox{red!39}{\strut few} \colorbox{red!47}{\strut UNK} \colorbox{red!33}{\strut in} \colorbox{red!26}{\strut the} \colorbox{red!37}{\strut area} \colorbox{red!25}{\strut ,} \colorbox{red!23}{\strut but} \colorbox{red!21}{\strut i} \colorbox{red!30}{\strut 'm} \colorbox{red!31}{\strut not} \colorbox{red!9.8}{\strut sure} \colorbox{red!27}{\strut how} \colorbox{red!20}{\strut to} \colorbox{red!62}{\strut find} \colorbox{red!61}{\strut them} \colorbox{red!13}{\strut .} \colorbox{red!66}{\strut http} \colorbox{red!0}{\strut :} \colorbox{red!0.2}{\strut UNK} \colorbox{red!18}{\strut $<$/s$>$}
} \\
\midrule
\textsf{\tiny 
\colorbox{cyan!4.9}{\strut $<$s$>$} \colorbox{cyan!27}{\strut which} \colorbox{cyan!46}{\strut country} \colorbox{cyan!5.8}{\strut is} \colorbox{cyan!0}{\strut the} \colorbox{cyan!40}{\strut best} \colorbox{cyan!14}{\strut at} \colorbox{cyan!52}{\strut soccer} \colorbox{cyan!9.8}{\strut ?} \colorbox{cyan!95}{\strut brazil} \colorbox{cyan!7.1}{\strut or} \colorbox{cyan!43}{\strut germany} \colorbox{cyan!11}{\strut .} \colorbox{cyan!23}{\strut $<$/s$>$}
} \\
\vspace{0.25mm}
\textsf{\tiny 
\colorbox{red!54}{\strut who} \colorbox{red!18}{\strut is} \colorbox{red!26}{\strut the} \colorbox{red!98}{\strut best} \colorbox{red!64}{\strut soccer} \colorbox{red!0.73}{\strut player} \colorbox{red!8.6}{\strut in} \colorbox{red!24}{\strut the} \colorbox{red!0}{\strut world} \colorbox{red!0.55}{\strut ?} \colorbox{red!7.7}{\strut i} \colorbox{red!14}{\strut think} \colorbox{red!47}{\strut he} \colorbox{red!12}{\strut is} \colorbox{red!19}{\strut the} \colorbox{red!26}{\strut best} \colorbox{red!36}{\strut player} \colorbox{red!6.2}{\strut in} \colorbox{red!21}{\strut the} \colorbox{red!25}{\strut world} \colorbox{red!11}{\strut .} \colorbox{red!64}{\strut ronaldinho} \colorbox{red!11}{\strut is} \colorbox{red!5.8}{\strut the} \colorbox{red!35}{\strut best} \colorbox{red!9.9}{\strut player} \colorbox{red!4.5}{\strut in} \colorbox{red!0.54}{\strut the} \colorbox{red!6.5}{\strut world} \colorbox{red!6.6}{\strut .} \colorbox{red!28}{\strut he} \colorbox{red!8.2}{\strut is} \colorbox{red!13}{\strut a} \colorbox{red!27}{\strut great} \colorbox{red!6.9}{\strut player} \colorbox{red!12}{\strut .} \colorbox{red!44}{\strut $<$/s$>$}
}\\
\vspace{0.25mm}
\textsf{\tiny 
\colorbox{red!44}{\strut will} \colorbox{red!99}{\strut ghana} \colorbox{red!8.5}{\strut be} \colorbox{red!40}{\strut able} \colorbox{red!0}{\strut to} \colorbox{red!39}{\strut play} \colorbox{red!27}{\strut the} \colorbox{red!41}{\strut next} \colorbox{red!52}{\strut game} \colorbox{red!34}{\strut in} \colorbox{red!62}{\strut 2010} \colorbox{red!83}{\strut fifa} \colorbox{red!20}{\strut world} \colorbox{red!2.3}{\strut cup} \colorbox{red!35}{\strut ?} \colorbox{red!26}{\strut yes} \colorbox{red!6.6}{\strut ,} \colorbox{red!48}{\strut they} \colorbox{red!8.4}{\strut will} \colorbox{red!33}{\strut win} \colorbox{red!18}{\strut it} \colorbox{red!25}{\strut all} \colorbox{red!22}{\strut .} \colorbox{red!6.4}{\strut $<$/s$>$}
}\\
\bottomrule
\end{tabularx}
\vspace{-3mm}
\caption{(Top) Saliency visualization of some examples from the test set. Here the saliency values are rescaled to be between 0-100 within each example for easier visualization. Red indicates higher saliency values. (Middle) Input saliency of the first test example from the top (in blue), in addition to two sample outputs generated from the variational posterior (with their saliency values in red). (Bottom) Same as the middle except we use a made-up example. Best viewed in color.}
\label{fig:heatmap}
\vspace{-5.5mm}
\end{figure*}
We visualize the saliency of a few examples from the test set in Figure~\ref{fig:heatmap} (top). Each example consists of a question followed by an answer from the Yahoo corpus. From a qualitative analysis several things are apparent: the latent variable seems to encode question type (i.e. if, what, how, why, etc.) and therefore saliency is high for the first word; content words (nouns, adjectives, lexical verbs) have much higher saliency than function words (determiners, prepositions, conjunctions, etc.); saliency of the \textsf{\small $<$/s$>$} token is quite high, indicating that the length information is also encoded in the latent space. In the third example we observe that the left parenthesis has higher saliency than the right parenthesis (0.32 vs. 0.24 on average across the test set), as the latter can be predicted by conditioning on the former rather than on the latent representation $\zvec$.

The previous definition of saliency measures the influence of $\zvec$ on the \emph{output} $\xvec_t$. We can 
also roughly measure the influence of the \emph{input} $\xvec_t$ on the latent representation $\zvec$, which we
refer to as input saliency:
\[  \Big\Vert \,  \E_{q(\zvec^{} \param \lambda^{})}\Big[ \frac{\diff \Vert \zvec \Vert_2}{\diff \mathbf{w}_t} \Big] \, \Big\Vert_2 \] 
Here $\mathbf{w}_t$ is the encoder word embedding for $\xvec_t$.\footnote{As the norm of $\zvec$ is a rather crude measure, a better measure would be obtained by analyzing the spectra of the Jacobian $\frac{\diff \zvec}{\diff \mathbf{w}_t}$. However this is computationally too expensive to calculate for each token in the corpus. }
We visualize the input saliency for a test example (Figure~\ref{fig:heatmap}, middle) and a made-up example (Figure~\ref{fig:heatmap}, bottom). Under each input example we also visualize a two samples from the variational posterior, and find that the generated examples are often meaningfully related to the input example.\footnote{We first sample $\zvec \sim q(\zvec \param \lambda_K)$ then $\xvec \sim p(\xvec \given \zvec \param \theta)$. When sampling $\xvec_t \sim p(\xvec_t \given \xvec_{<t}, \zvec)$ we sample with temperature $T = 0.25$, i.e. $p(\xvec_t \given \xvec_{<t}, \zvec) = \softmax(\frac{1}{T} \mathbf{s}_{t})$ where $\mathbf{s}_{t}$ is the vector with scores for all words. We found the generated examples to be related to the original (in some way) in roughly half the cases.}

\begin{figure*}[t]
\vspace{-3mm}
\center
\begin{subfigure}{0.24\textwidth}
\center
\includegraphics[scale=0.23]{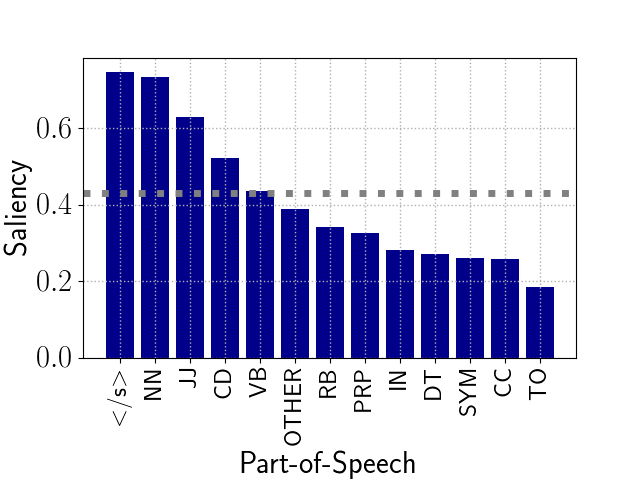}
\end{subfigure}
\begin{subfigure}{0.24\textwidth}
\center
\includegraphics[scale=0.23]{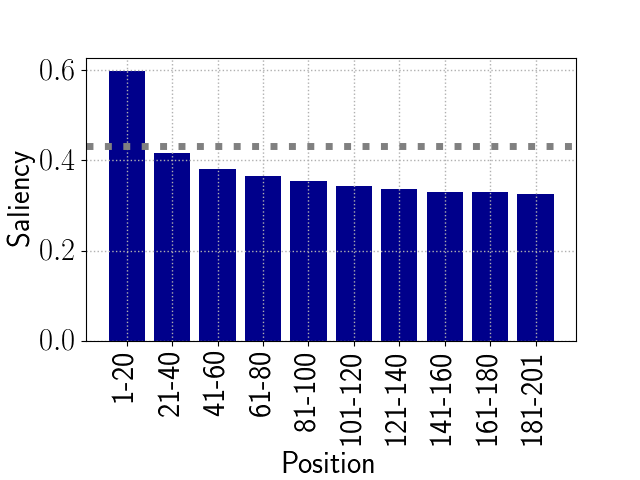}
\end{subfigure}
\begin{subfigure}{0.24\textwidth}
\center
\includegraphics[scale=0.23]{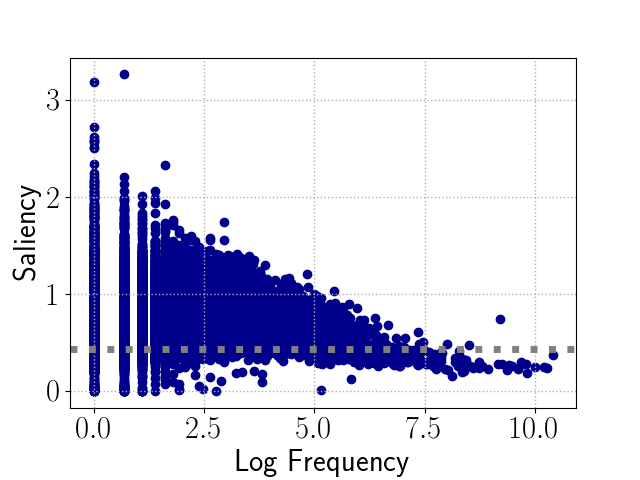}
\end{subfigure} 
\begin{subfigure}{0.24\textwidth}
\center
\includegraphics[scale=0.23]{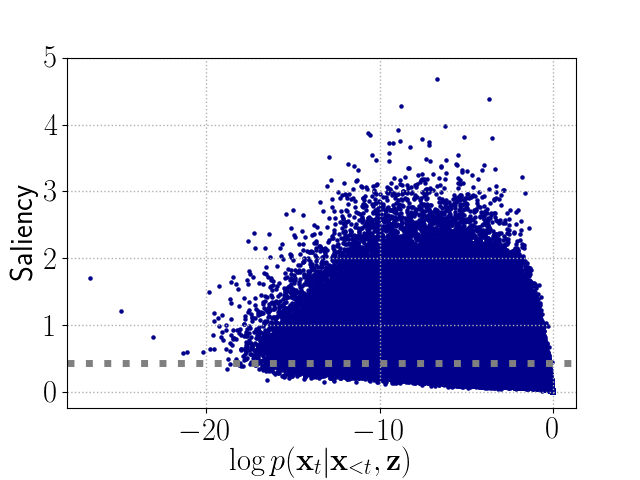}
\end{subfigure} 
\vspace{-3mm}
\caption{Output saliency by part-of-speech tag, position, log frequency, and log-likelihood. See Section~\ref{sec:latent} for the definitions of output saliency. The dotted gray line in each plot shows the average saliency across all words.}
\label{fig:sal}
\vspace{-5mm}
\end{figure*}

We quantitatively analyze output saliency across part-of-speech, token position, word frequency, and log-likelihood in Figure~\ref{fig:sal}: nouns (NN), adjectives (JJ), verbs (VB), numbers (CD), and the \textsf{\small $<$/s$>$} token have higher saliency than conjunctions (CC), determiners (DT), prepositions (IN), and the TO token---the latter are relatively easier to predict by conditioning on previous tokens; similarly, on average, tokens occurring earlier have much higher saliency than those occurring later (Figure~\ref{fig:sal} shows absolute position but the plot is similar with relative position); the latent variable is used much more when predicting rare tokens; there is some negative correlation between saliency and log-likelihood (-0.51), though this relationship does not always hold---e.g. \textsf{\small $<$/s$>$} has high saliency but is relatively easy to predict with an average log-likelihood of -1.61 (vs. average log-likelihood of -4.10 for all tokens). Appendix D has the corresponding analysis for input saliency, which are qualitatively similar. 

These results seem to suggest that the latent variables are encoding interesting and potentially interpretable aspects of language. While left as future work, it is possible that manipulations in the latent space of a model learned this way could lead to controlled generation/manipulation of output text \cite{Hu2017,Mueller2017}.
\vspace{-3mm}
\subsection{Limitations}
\vspace{-2mm}
A drawback of our approach (and other non-amortized inference methods) is that each training step requires backpropagating through the generative model multiple times, which can be costly especially if the generative model is expensive to compute (e.g. LSTM/PixelCNN). This may potentially be mitigated through more sophisticated \textit{meta learning} approaches \cite{Andrychowicz2016,Marino2017}, or with more efficient use of the past gradient information during SVI via averaging \cite{Schmidt2013} or importance sampling \cite{Sakaya2017}.
One could also consider employing synthetic gradients \cite{Jaderberg2017} to limit the number of backpropagation steps during training. \citet{Krishnan2017} observe that it is more important to train with iterative refinement during earlier stages (we also observed this in preliminary experiments), and therefore annealing the number of refinement steps as training progresses could also speed up training.

Our approach is mainly applicable to variational families that avail themselves to differentiable optimization (e.g. gradient ascent) with respect to the ELBO, which include much recent work on employing more flexible variational families with VAEs. In contrast, VAE + SVI and VAE + SVI + KL are applicable to more general optimization algorithms. 
\vspace{-3mm}
\section{Related Work}
\vspace{-2mm}
Our work is most closely related the line of work which uses a separate model to initialize variational parameters and subsequently updates them through an iterative procedure \cite{Salakhutdinov2010,Cho2013,Salimans2015,Hjelm2016,Krishnan2017,Pu2017}. 
\citet{Marino2017} utilize meta-learning to train an inference network which learns to perform iterative inference by training a deep model to output the variational parameters for each time step.

While differentiating through inference/optimization was initially explored by various researchers primarily outside the area of deep learning \cite{Stoyanov2011,Domke2012,Brakel2013}, they have more recently been explored in the context of hyperparameter optimization \cite{Maclaurin2015} and as a differentiable layer of a deep model \cite{Belanger2017,Kim2017,Metz2017,Amos2017}. 

Initial work on VAE-based approaches to image modeling focused on simple generative models that assumed independence among pixels conditioned on the latent variable \cite{Kingma2014,Rezende2014}. More recent works have obtained substantial improvements in log-likelihood and sample quality through utilizing powerful autoregressive models (PixelCNN) as the generative model \cite{Chen2017,Gulrajani2017}. 

In contrast, modeling text with VAEs has remained challenging. \citet{Bowman2016} found that
using an LSTM generative model resulted in a degenerate case whereby the variational posterior collapsed to the prior and the generative model ignored the latent code (even with richer variational families). Many works on VAEs for text have thus made simplifying conditional independence assumptions \cite{Miao2016,Miao2017}, used less powerful generative models such as convolutional networks \cite{Yang2017,Semeniuta2017}, or combined a recurrent generative model with a topic model \cite{Dieng2017,Wang2018}. Note that unlike to sequential VAEs that employ different latent variables at each time step \cite{Chung2015,Fraccaro2016,Krishnan2017b,Serban2017,Goyal2017b}, in this work we focus on modeling the entire sequence with a global latent variable.

Finally, since our work only addresses the \textit{amortization gap} (the gap between the log-likelihood and the ELBO due to amortization) and not the \textit{approximation gap} (due to the choice of a particular variational family) \cite{Cremer2017}, it can be combined with existing work on employing richer posterior/prior distributions within the VAE framework \cite{Rezende2015,Kingma2016,Johnson2016,Tran2016,Goyal2017,Guu2017,Tomczak2017}.
\vspace{-3mm}
\section{Conclusion}
\vspace{-2mm}
This work outlines semi-amortized variational autoencoders, which combine amortized inference with local iterative refinement to train deep generative models of text and images. With the approach we find that we are able to train deep latent variable models of text with an expressive autogressive generative model that does not ignore the latent code.

From the perspective of learning latent representations, one might question the prudence of using an autoregressive model that fully conditions on its entire history (as opposed to assuming some conditional independence) given that $p(\xvec)$ can always be factorized as $\prod_{t=1}^T p(\xvec_t \given \xvec_{<t})$, and therefore the model is \emph{non-identifiable} (i.e. it does not have to utilize the latent variable). However in finite data regimes we might still expect a model that makes use of its latent variable to generalize better due to potentially better inductive bias (from the latent variable). Training generative models that both model the underlying data well and learn good latent representations is an important avenue for future work.

\newpage
\section*{Acknowledgements}
\vspace{-2mm}
{\small
We thank Rahul Krishnan, Rachit Singh, and Justin Chiu for insightful comments/discussion. We additionally thank Zichao Yang for providing the text dataset.
YK and AM are supported by Samsung Research. SW is supported by an Amazon AWS ML Award.}
\vspace{-6mm}
{\footnotesize
\bibliography{note}
\bibliographystyle{icml2018}
}

\newpage
\twocolumn[
\icmltitle{Supplementary Materials for Semi-Amortized Variational Autoencoders}
]
\appendix
\section{Training Semi-Amortized Variational Autoencoders with Gradient Clipping}
For stable training we found it crucial to modify Algorithm 1 to clip the gradients at various stages. This is shown in Algorithm 2, where we have a clipping parameter $\eta$.
The $\clip(\cdot)$ function is given by
\[
    \clip(\uvec, \eta)= 
\begin{cases}
    \frac{\eta}{\Vert \uvec \Vert_2}\uvec \,\, , & \text{if } \Vert \uvec \Vert_2 > \eta \\
    \uvec \,\, ,              & \text{otherwise}
\end{cases}
\]
We use $\eta = 5$ in all experiments. The finite difference estimation itself also uses gradient clipping. See \url{https://github.com/harvardnlp/sa-vae/blob/master/optim_n2n.py} for the exact implementation.

\section{Experimental Details}
For all the variational models we use a spherical Gaussian prior. The variational family is the diagonal Gaussian parameterized by the vector of means and log variances. For models trained with SVI the initial variational parameters are randomly initialized from a Gaussian with standard deviation equal to 0.1.
\subsection{Synthetic Data}
We generate synthetic data points according to the following generative process:
\begin{align*}
z_1, z_2 &\sim \mcN (0, 1) \,\,\,\,\,\,\,\,\,\,\,\,\,\, \hvec_t = \LSTM(\hvec_{t-1}, \xvec_t)  \\
&\xvec_{t+1} \sim \softmax(\MLP([\hvec_t , z_1, z_2]))
\end{align*}
Here LSTM is a one-layer LSTM with 100 hidden units where the input embedding is also 100-dimensional. The initial hidden/cell states are set to zero, and we generate for 5 time steps for each example (so $\xvec = [\xvec_1, \dots, \xvec_5]$). The MLP consists of a single affine transformation to project out to the vocabulary space, which has 1000 tokens. LSTM/MLP parameters are randomly initialized with $\mcU(-1, 1)$, except for the part of the MLP that directly connects to the latent variables, which is initialized with $\mcU(-5, 5)$. This is done to make sure that the latent variables have more influence in predicting $\xvec$. We generate 5000 training/validation/test examples.

When we learn the generative model the LSTM is initialized over $\mcU(-0.1, 0.1)$. The inference network is also a one-layer LSTM with 100-dimensional hidden units/input embeddings, where the variational parameters are predicted via an affine transformation on the final hidden state of the encoder. All models are trained with stochastic gradient descent with batch size 50, learning rate 1.0, and gradient clipping at 5. The learning rate starts decaying by a factor of 2 each epoch after the first epoch at which validation performance does not improve. This learning rate decay is not triggered for the first 5 epochs. We train for 20 epochs, which was enough for convergence of all models.
For SVI/SA-VAE we perform 20 steps of iterative inference with stochastic gradient descent and learning rate 1.0 with gradient clipping at 5.

\begin{algorithm}[tb]
   \caption{Semi-Amortized Variational Autoencoders with Gradient Clipping}
   \label{alg:savi2}
\begin{algorithmic}
   \STATE {\bfseries Input:} inference network $\phi$, generative model $\theta$, \\ 
   \hspace{10mm} inference steps $K$, learning rate $\alpha$, momentum $\gamma$,   \\
   \hspace{10mm} loss function $f(\lambda, \theta, \xvec) = -\ELBO(\lambda, \theta, \xvec)$, \\
   \hspace{10mm} gradient clipping parameter $\eta$

   \STATE Sample $\xvec^{} \sim p_\mathcal{D}(\xvec)$
   \STATE $\lambda_0 \gets \enc(\xvec^{} \param \phi)$ 
   \STATE $v_0 \gets 0$
   \FOR{$k=0$ {\bfseries to} $K-1$} 
   \STATE $v_{k+1} \gets \gamma v_{k} - \clip(\nabla_{\lambda}f(\lambda_k, \theta, \xvec^{}), \eta)$
   \STATE $\lambda_{k+1} \gets \lambda_k + \alpha v_{k+1}$
   \ENDFOR
   \STATE  $\mcL \gets f(\lambda_K, \theta, \xvec^{})$ 
   \STATE $\overline{\lambda}_K \gets \nabla_{\lambda}f(\lambda_K, \theta, \xvec^{}) $
   \STATE $\overline{\theta} \gets \nabla_{\theta}f(\lambda_K, \theta, \xvec^{}) $ 
   \STATE $\overline{v}_K \gets 0$
   \FOR{$k=K-1$ {\bfseries to} $0$}
   \STATE $\overline{v}_{k+1} \gets \overline{v}_{k+1} + \alpha \overline{\lambda}_{k+1}$
   \STATE $\overline{\lambda}_k \gets \overline{\lambda}_{k+1} - \Hess_{\lambda, \lambda}f(\lambda_k, \theta, \xvec^{}) \overline{v}_{k+1} $
   \STATE $\overline{\lambda}_k \gets \clip(\overline{\lambda}_k , \eta)$
   \STATE $\overline{\theta} \gets \overline{\theta} - \clip(\Hess_{\theta,\lambda}f(\lambda_k, \theta, \xvec^{}) \overline{v}_{k+1}, \eta)$
    \STATE $\overline{v}_{k} \gets \gamma \overline{v}_{k+1}$
   \ENDFOR
   \STATE $\frac{\diff \mcL}{\diff \theta} \gets \overline{\theta} $   
   \STATE $\frac{\diff \mcL}{\diff \phi} \gets \frac{\diff \lambda_0}{\diff \phi} \overline{\lambda}_0 $ 
   \STATE Update $\theta, \phi$ based on $\frac{\diff \mcL}{\diff \theta}, \frac{\diff \mcL}{\diff \phi}$
 
\end{algorithmic}
\end{algorithm}

\begin{table*}
\centering
\begin{sc}
\begin{small}
\begin{tabular}{l c c c c c }
\toprule
Inference Network & \multicolumn{2}{c}{3-layer ResNet} & & \multicolumn{2}{c}{2-layer MLP}   \\
\midrule
Model & VAE & SA-VAE & & VAE & SA-VAE \\
\midrule
Data Size: 25\%& $  92.21 \,\, (0.81)$ & $91.89 \,\,(2.31)$  & & $92.33 \,\, (0.27)$ & $92.03 \,\, (1.32)$ \\
Data Size: 50\% & $  91.38 \,\, (0.77)$ & $91.01 \,\,(2.54)$  & & $91.40 \,\, (0.51)$ & $91.10 \,\, (1.48)$ \\
Data Size: 75\% & $  90.82 \,\, (1.06)$ & $90.51 \,\,(2.07)$  & &$90.90 \,\, (0.45)$ & $90.67 \,\, (1.34)$ \\
Data Size: 100\%& $  90.43 \,\, (0.98)$ & $90.05 \,\,(2.78)$  & & $90.56 \,\, (0.61)$ & $90.25 \,\, (1.77)$ \\
\midrule
1-layer PixelCNN &$  96.53 \,\, (10.36)$ & $96.01 \,\,(10.93)$  & & $98.30 \,\, (8.87)$ & $96.43 \,\, (10.14)$ \\
3-layer PixelCNN  &$  93.75 \,\, (7.10)$ & $93.16 \,\,(8.73)$  & & $94.45 \,\, (5.46)$ & $93.55 \,\, (7.20)$ \\
6-layer PixelCNN &$  91.24 \,\, (3.25)$ & $90.79 \,\,(4.44)$  &  & $91.40 \,\, (2.06)$ & $91.01 \,\, (3.27)$ \\
9-layer PixelCNN &$  90.54 \,\, (1.78)$ & $90.28 \,\,(3.02)$  & & $90.72 \,\, (1.14)$ & $90.34 \,\, (2.26)$ \\
12-layer PixelCNN &$  90.43 \,\, (0.98)$ & $90.05 \,\,(2.78)$  & & $90.56 \,\, (0.61)$ & $90.25 \,\, (1.77)$ \\
\bottomrule
\end{tabular}
\end{small}
\end{sc}
\vspace{-1mm}
\caption{Upper bounds on negative log-likelihood (i.e. negative ELBO) of VAE/SA-VAE trained on OMNIGLOT, where   we vary the capacity of the inference network (3-layer ResNet vs 2-layer MLP). KL portion of the loss is shown in parentheses. (Top) Here we vary the training set size from 25\% to 100\%, and use a 12-layer PixelCNN as the generative model. (Bottom) Here we fix the training set size to be 100\%, and vary the capacity of the generative model.}
\label{tab:omni2}
\end{table*}

\subsection{Text}
We use the same model architecture as was used in \citet{Yang2017}.
The inference network and the generative model are both one-layer LSTMs with 1024-dimensional hidden states where the input word embedding is 512-dimensional. We use the final hidden state of the encoder
to predict (via an affine transformation) the vector of variational means and log variances. The latent space is 32-dimensional. 
The sample from the variational posterior is used to initialize the initial hidden state of the generative LSTM (but not the cell state) via an affine transformation, and additionally fed as input (i.e. concatenated with the word embedding) at each time step. There are dropout layers with probability 0.5 between the input-to-hidden layer and the hidden-to-output layer on the generative LSTM only.

The data contains 100000/10000/10000 train/validation/test examples with 20000 words in the vocabulary. All models are trained with stochastic gradient descent with batch size 32 and learning rate 1.0, where the learning rate starts decaying by a factor of 2 each epoch after the first epoch at which validation performance does not improve. This learning rate decay is not triggered for the first 15 epochs to ensure adequate training. We train for 30 epochs or until the learning rate has decayed 5 times, which was enough for convergence for all models. Model parameters are initialized over $\mcU(-0.1, 0.1)$ and gradients are clipped at 5. We employ a KL-cost annealing schedule whereby the multiplier on the KL-cost term is increased linearly from 0.1 to 1.0 each batch over 10 epochs.
For models trained with iterative inference we perform SVI via stochastic gradient descent with momentum 0.5 and learning rate 1.0. Gradients are clipped after each step of SVI (also at 5).

\subsection{Images}
\begin{figure*}[t]
\center
\begin{subfigure}{0.33\textwidth}
\center
\includegraphics[scale=0.3]{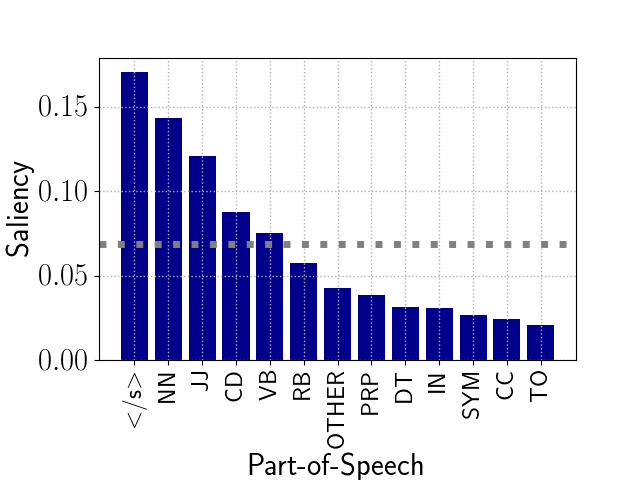}
\end{subfigure}
\begin{subfigure}{0.33\textwidth}
\center
\includegraphics[scale=0.3]{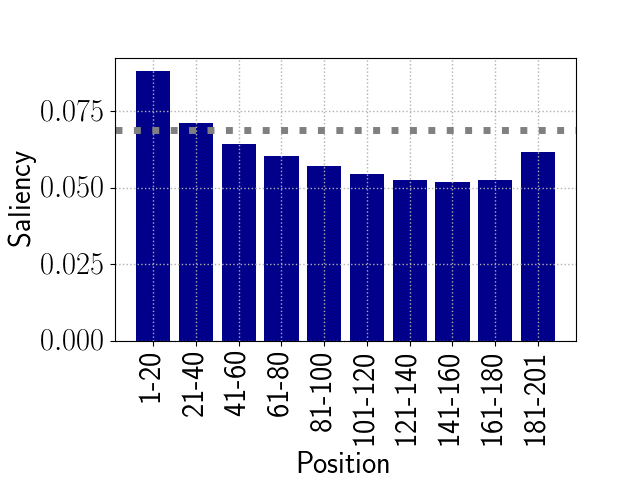}
\end{subfigure}
\begin{subfigure}{0.33\textwidth}
\center
\includegraphics[scale=0.3]{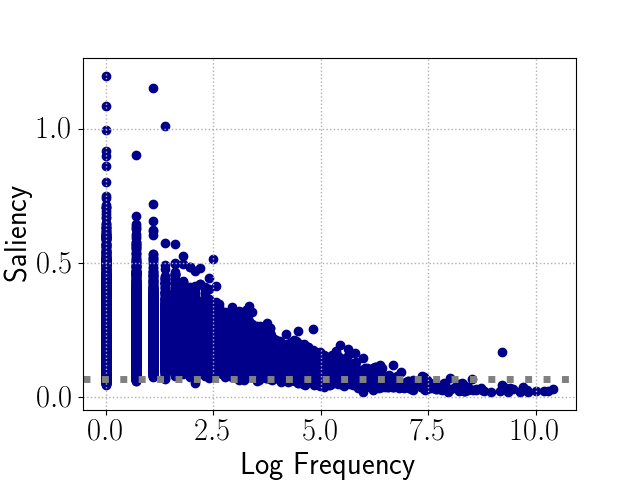}
\label{fig:sal2}
\end{subfigure}
\caption{Input saliency by part-of-speech tag (left), position (center), and log frequency (right). The dotted gray line in each plot shows the average saliency across all words.}
\end{figure*}

The preprocessed OMNIGLOT dataset does not have a standard validation split so we randomly pick 2000 images from training as validation. As with previous works the pixel value is scaled to be between 0 and 1 and interpreted as probabilities, and the images are dynamically binarized during training.

Our inference network consists of 3 residual blocks where each block is made up of a standard residual layer (i.e. two convolutional layers with $3\times 3$ filters, ReLU nonlinearities, batch normalization, and residual connections) followed by a downsampling convolutional layer with filter size and stride equal to 2. These layers have 64 feature maps. The output of residual network is flattened and then used to obtain the variational means/log variances via an affine transformation.

The sample from the variational distribution (which is 32-dimensional) is first projected out to the image spatial resolution with 4 feature maps (i.e. $4 \times 28 \times 28$) via a linear transformation, then concatenated with the original image, and finally fed as input to a 12-layer Gated PixelCNN \cite{Oord2016b}.
The PixelCNN has three $9 \times 9$ layers, followed by three $7 \times 7$ layers, then three $5 \times 5$ layers, and finally three $3 \times 3$ layers. All the layers have 32 feature maps, and there is a final $1\times 1$ convolutional layer followed by a sigmoid nonlinearity to produce a distribution over binary output. The layers are appropriately masked to ensure that the distribution over each pixel is conditioned only on the pixels left/top of it. We train with Adam with learning rate 0.001, $\beta_1$ = 0.9, $\beta_2$ = 0.999 for 100 epochs with batch size of 50. Gradients are clipped at 5. 

For models trained with iterative inference we perform SVI via stochastic gradient descent with momentum 0.5 and learning rate 1.0, with gradient clipping (also at 5).

\section{Data Size/Model Capacity}
In Table 4 we investigate the performance of VAE/SA-VAE as we vary the capacity of the inference network, size of the training set, and the capacity of the generative model. The MLP inference network has two ReLU layers with 128 hidden units. For varying the PixelCNN generative model, we sequentially remove layers from our baseline 12-layer model starting from the bottom (so the 9-layer PixelCNN has three $7 \times 7$ layers, three $5 \times 5$ layers, three $3 \times 3$ layers, all with 32 feature maps).

Intuitively, we expect iterative inference to help more when the inference network and the generative model are less powerful, and we indeed see this in Table~\ref{tab:omni2}. Further, one might expect SA-VAE to be more helpful in  small-data regimes as it is harder for the inference network amortize inference and generalize well to unseen data.
However we find that SA-VAE outperforms VAE by a similar margin across all training set sizes.

Finally, we observe that across all scenarios the KL portion of the loss is much higher for models trained with SA-VAE, indicating that these models are learning generative models that make more use of the latent representations.

\section{Input Saliency Analysis}

In Figure 5 we show the input saliency by part-of-speech tag (left), position (center), and frequency (right).
Input saliency of a token $\xvec_t$ is defined as:
\[  \Big\Vert \,  \E_{q(\zvec^{} \param \lambda^{})}\Big[ \frac{\diff \Vert \zvec \Vert_2}{\diff \mathbf{w}_t} \Big] \, \Big\Vert_2 \]
Here $\mathbf{w}_t$ is the encoder word embedding for $\xvec_t$. Part-of-speech tagging is done using NLTK.
\end{document}